\documentclass[10pt,twocolumn,letterpaper]{article}
\usepackage[pagenumbers]{cvpr} 

\usepackage{graphicx}
\usepackage{amsmath}
\usepackage{amssymb}
\usepackage{stackengine}
\usepackage{booktabs}
\usepackage{multirow}
\usepackage{multicol}
\usepackage{graphicx,mathtools,kantlipsum}
\newcommand\xR{$\mathit{x}$R}
\usepackage{tabularx} 
\newcommand{\figref}[1]{Fig.~\ref{#1}}

\newcommand{\tabref}[1]{Tab.~\ref{#1}}
\usepackage{xcolor}
\newcommand{\edit}[1]{\textcolor{black}{#1}}

\usepackage[pagebackref,breaklinks,colorlinks]{hyperref}

\usepackage[capitalize]{cleveref}
\crefname{section}{Sec.}{Secs.}
\Crefname{section}{Section}{Sections}
\Crefname{table}{Table}{Tables}
\crefname{table}{Tab.}{Tabs.}

\def\cvprPaperID{*****} 
\def\confName{CVPR}
\def\confYear{2022}

\usepackage{placeins}

\begin{document}

\title{Building Spatio-temporal Transformers for Egocentric 3D Pose Estimation}

\author{Jinman Park$^\dagger$, Kimathi Kaai$^\dagger$, Saad Hossain$^\dagger$, Norikatsu Sumi$^\#$, Sirisha Rambhatla$^\dagger$, and Paul Fieguth$^\dagger$ \\
$^\dagger$University of Waterloo, Waterloo, ON, Canada\\
$^\#$Mobility and AI Lab, Nissan Motor Co. Ltd., Atsugi-shi, Kanagawa, Japan\\
{\tt\footnotesize $^\dagger$\{j97park, kkaai, s42hossa, srambhat, pfieguth\}@uwaterloo.ca}, {\tt\footnotesize $^\#$norikatsu-sumi@mail.nissan.co.jp}
\vspace*{-25pt}}

\twocolumn[{
\vspace*{-15pt}
\maketitle

\renewcommand\twocolumn[1][]{#1}


\begin{center}
    \centering
    \captionsetup{type=figure}
    \includegraphics[width=0.75\textwidth]{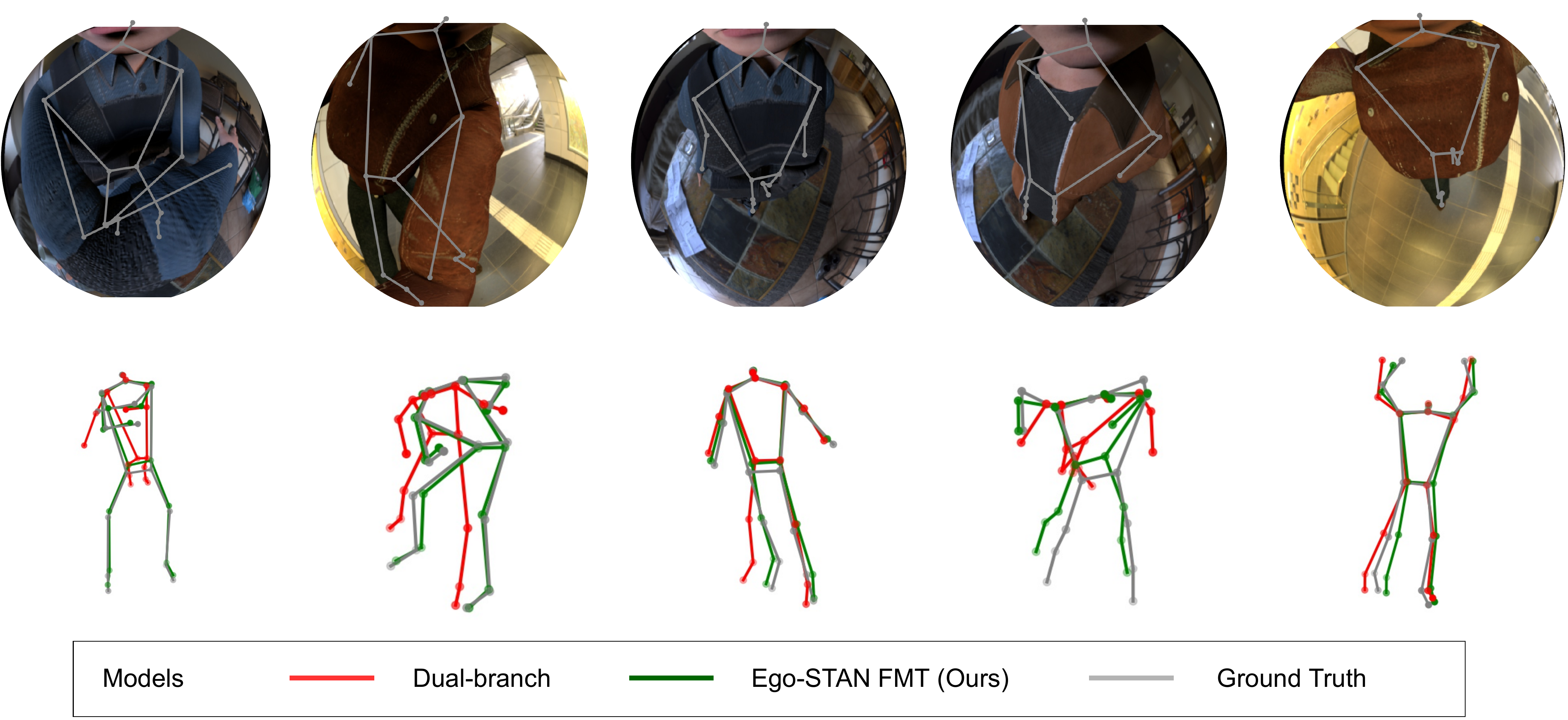}
    \captionof{figure}{\edit{\textbf{Qualitative evaluation of our proposed method on frames with high self-occlusion}. The images compare the performance of a feature map token-based variant of our proposed method --- Ego-STAN FMT with state-of-the-art dual-branch authoencoder-based model \cite{tome2019xr} on self-occluded frames. We observe that Ego-STAN exhibits superior performance as compared to the SOTA \cite{tome2019xr}.}}
    \label{fig:qualitative}
\end{center}
}]

\begin{abstract}
\vspace{-10pt}
 \edit{Egocentric 3D human pose estimation (HPE) from images is challenging due to severe self-occlusions and strong distortion introduced by the fish-eye view from the head mounted camera. Although existing works use intermediate heatmap-based representations to counter distortion with some success, addressing self-occlusion remains an open problem. In this work, we leverage information from past frames to guide our self-attention-based 3D HPE estimation procedure --- Ego-STAN. Specifically, we build a spatio-temporal Transformer model that attends to semantically rich convolutional neural network-based feature maps. We also propose feature map tokens: a new set of learnable parameters to attend to these feature maps. Finally, we demonstrate Ego-STAN's superior performance on the \xR{-EgoPose} dataset where it achieves a 30.6\% improvement on the overall mean per-joint position error, while leading to a 22\% drop in parameters compared to the state-of-the-art.}
\end{abstract}

\vspace{-10pt}
\section{Introduction}
\label{sec:intro}
\edit{Virtual immersive technologies, such as augmented, virtual, and mixed reality environments (\xR{}), which rely on user-centric customizations and viewpoint rendering, have underscored the need for accurate human pose estimation (HPE) to support a wide range of applications including medical training \cite{vaughan2020scoring} and architecture \cite{grepon2021architectural}, among others. 
Although accurate 3D HPE can be performed using motion capture systems by attaching sensors to major human joints, this approach may not be suitable for real-world settings, due to the cost and additional infrastructural needs \cite{h36m_pami}. To this end, image-based 3D HPE has emerged as a practical alternative \cite{li2015maximum, sun2017compositional, tekin2016structured, pavlakos2018ordinal}.}
\begin{figure*}
\centerline{\includegraphics[width=0.9\textwidth]{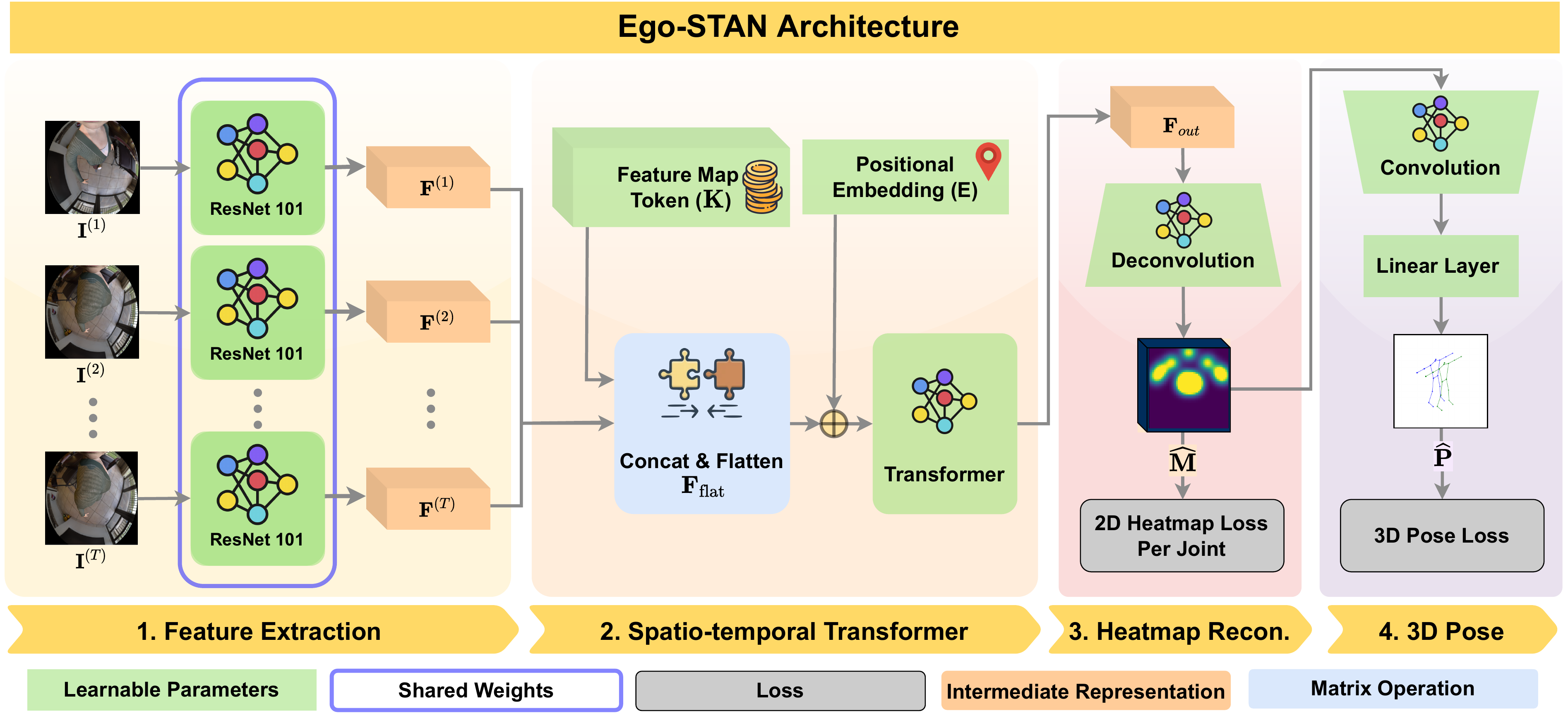}}
\caption{\textbf{Ego-STAN Overview}. The proposed Ego-STAN model captures the dynamics of human motion in ego-centric images using Transformer-based spatio-temporal modeling. Our Transformer architecture leverages \emph{feature map tokens} to facilitate spatio-temporal attention to semantically rich feature maps. Our heatmap reconstruction module estimates the 2D heatmap using deconvolutions, which are used by the 3D pose estimator to estimate the 3D joint coordinates.}
\label{fig:Ego-STAN}
\vspace{-12pt}
\end{figure*}

\edit{The image-based 3D HPE literature is primarily devoted to settings where a static camera observes an entire scene \cite{chen20173d, martinez2017simple, zhou2019hemlets}, however this modality is not ideal for applications requiring higher and robust (low variance) accuracies. 
The egocentric modality offers some respite allowing mobility and the flexibility to focus on the subject even in cluttered environments \cite{tome2019xr, xu2019mo, wang2021estimating}. Notwithstanding these advantages, egocentric views introduce challenges of \emph{distortion} (e.g. lower body joints are visually much smaller than the upper body joints) and self-occlusion (e.g. lower body joints heavily occluded by the upper torso); see \figref{fig:qualitative}.}

\edit{
Recent works on egocentric HPE propose using a dual-branch autoencoder-based 2D heatmap to 3D pose estimator \cite{tome2019xr}, and extra camera information \cite{9423180} from static egocentric images. However, self-occlusions are challenging to address from these images-only approaches. 
}

\edit{The question we aim to address is: \textit{how can we accurately estimate 3D human pose from egocentric images while jointly addressing the distortion and self-occlusions caused by these views}? Motivated from recent works for static cameras, which leverage spatio-temporal modeling to improve HPE \cite{zheng20213d}, we build Egocentric Spatio-Temporal Self-Attention Network (Ego-STAN) which leverages \emph{feature map tokens} (FMT), a specialized spatio-temporal attention, heatmap-based representations, and 2D heatmap to 3D HPE module.  Ego-STAN achieves an \textbf{overall improvement of 30.6\%} mean per-joint position error (MPJPE) compared to the SOTA \cite{tome2019xr}, while \textbf{leading to a 22\% reduction in the trainable parameters} on the $x$R-EgoPose dataset \cite{tome2019xr}; see \figref{fig:qualitative} for a comparative analysis. }

\vspace{-15pt}

\section{Related Work}
The Mo2Cap2 dataset was one of the first large egocentric HPE synthetic datasets \cite{xu2019mo}, however it is not amenable to spatio-temporal modeling since it only consists of static images. The xR-EgoPose dataset \cite{tome2019xr}, which offers egocentric image sequences, was introduced to mitigate issues encountered in Mo2Cap2 by improving the quality of synthetic images, reflecting more realistic settings. The authors also introduced single and dual-branch auto-encoder structure(s) based on 2D heatmap and 3D pose reconstruction, following which \cite{9423180} leverages extra camera parameters to mitigate the fish-eye distortion.

More recently, utilizing \cite{tome2019xr} as a submodule, GlobalPose \cite{wang2021estimating} developed a sequential variational auto-encoder-based model to address depth ambiguity and temporal instability in egocentric HPE. Ego-STAN can be used with such methods.
\section{Ego-STAN: Ego-centric Spatio-Temporal Self-Attention Network}

\edit{We now provide a brief overview of the proposed \textbf{Ego}centric \textbf{S}patio-\textbf{T}emporal Self-\textbf{A}ttention \textbf{N}etwork (Ego-STAN) model, which jointly address the  self-occlusion and the distortion introduced by the ego-centric views. 
Ego-STAN (shown in Fig.~\ref{fig:Ego-STAN}) consists of four modules. Of these, the goals of the \textbf{feature extraction} and \textbf{spatio-temporal Transformer} modules are to address the self-occlusion problem by aggregating information from multiple time steps, while the \textbf{heatmap reconstruction} and \textbf{3D pose estimator} modules aim to accomplish uncertainty saturation with lighter 2D heatmap-to-3D lifting architectures.}
\begingroup

\renewcommand{\arraystretch}{1.25}
\begin{table*}[t]
  \centering
  \caption{\edit{Quantitative evaluation against the popular outside-in 3D HPE work of Martinez \textit{et al.} \cite{martinez2017simple} and the SOTA egocentric 3D HPE work of Tome \textit{et al.} \cite{tome2019xr}. Our proposed Ego-STAN variations have the highest accuracy across nine actions with the feature map token (FMT) variant having the lowest overall MPJPE (lower is better). We report our results as the average over three different random seeds. Our proposed three variants have a very low standard deviation of 2.06, 0.04, and 0.07 for Avg, Slice and FMT respectively.}}
  \vspace{-2pt}
  \resizebox{\textwidth}{!}
  {\begin{tabular}{ll|cccccccccc}
  \toprule
    \textbf{Approach} & \textbf{\parbox{1.7cm}{Evaluation \\ error (mm)}} & \textbf{Game} & \textbf{Gest.} & \textbf{Greet} & \textbf{\parbox{1cm}{Lower \\ Stretch}} & \textbf{Pat} & \textbf{React} & \textbf{Talk} & \textbf{\parbox{1cm}{Upper \\ Stretch}} & \textbf{Walk} & \textbf{All}\\
  \midrule
  & Upper body & 58.5 & 66.7 & 54.8 & 70.0 & 59.3 & 77.8 & 54.1 & 89.7 & 74.1 & 79.4 \\
    Martinez \cite{martinez2017simple}& Lower body & 160.7 & 144.1 & 183.7 & 181.7 & 126.7 & 161.2 & 168.1 & 159.4 & 186.9 & 164.8\\
    & Average & 109.6 & 105.4 & 119.3 & 125.8 & 93.0 & 119.7 & 111.1 & 124.5 & 130.5 & 122.1\\
    \hline
    \multirow{3}{*}{\parbox{2.0cm}{Tome \cite{tome2019xr} \\ single-branch}}& Upper body & 114.4 & 106.7 & 99.3 & 90.0 & 99.1 & 147.5 & 95.1 & 119.0 & 104.3 & 112.5 \\
    & Lower body & 162.2 & 110.2 & 101.2 & 175.6 & 136.6 & 203.6 & 91.9 & 139.9 & 159.0 & 148.3\\
    & Average & 138.3 & 108.5 & 100.3 & 133.3 & 117.8 & 175.6 & 93.5 & 129.0 & 131.9 & 130.4\\
    \hline
    \multirow{3}{*}{\parbox{2.0cm}{Tome \cite{tome2019xr} \\ dual-branch}}& Upper body & 48.8 & 50.0 & 43.0 & 36.8 & 48.6 & 56.4 & 42.8 & 49.3 & 43.2 & 50.5 \\
    & Lower body & 65.1 & 50.4 & 46.1 & 65.2 & 70.2 & 65.2 & 45.0 & 58.8 & 72.2 & 65.9\\
    & Average & 56.0 & 50.2 & 44.6 & 51.5 & 59.4 & 60.8 & 43.9 & 53.9 & 57.7 & 58.2\\
    \hline
     \multirow{3}{*}{\parbox{2.0cm}{\textbf{Ego-STAN} \\ \textbf{Slice (Ours)}}} & Upper body &  {27.2} &  {30.0} &  {36.3} &  {24.0} &  {21.3} &  {25.4} &   {25.3} &  {34.2} &  {25.5} &  {30.2} \\
     & Lower body &  {38.5} & \textbf{30.9} & \textbf{33.2} &  {54.5} & \textbf{32.1} &  {35.6} & \textbf{29.5} &  {64.0} & \textbf{55.9} &  {55.5}\\
    & Average &  {32.9} &  {30.4}	&  {34.8} &  {39.2} &	\textbf{26.7} &  {30.5} & \textbf{27.4} &  {49.1} & \textbf{40.7} &  {42.8} \\
    \hline
    \multirow{3}{*}{\parbox{2.0cm}{\textbf{Ego-STAN} \\ \textbf{Avg. (Ours)}}} & Upper body & \textbf{25.4} & \textbf{26.7} & \textbf{31.2} &  {25.9} & \textbf{20.7} & \textbf{23.3} & \textbf{23.9} &  {33.7} &  {26.7} &  {29.9} \\
     & Lower body &  \textbf{38.1} &  {32.7} &  {35.0} &  {54.7} &  {34.6} & \textbf{34.3} &  {31.2} &  {61.2} &  {57.2} &  {54.3}\\
    & Average & \textbf{31.7} & \textbf{29.7}	& \textbf{33.1} &  {40.3} &	 {27.7} & \textbf{28.8} &  {27.5} &  {47.4} &	 {42.0} &  {42.1} \\
    \hline
     \multirow{3}{*}{\parbox{2.0cm}{\textbf{Ego-STAN} \\ \textbf{FMT (Ours)}}} & Upper body &  {25.8} &  {28.7} &  {35.4} & \textbf{23.4} &  {22.6} &  {24.1} &  {25.9} & \textbf{30.9} & \textbf{25.2} & \textbf{28.2} \\
    & Lower body &  {40.3} &  {34.5} &  {38.3} & \textbf{54.4} &  {35.9} & {35.0} &  {33.4} & \textbf{57.6} &  {56.5} & \textbf{52.6}\\
    & Average &  {33.1} &  {31.6}	&  {36.9} & \textbf{38.9} &	 {29.2} &  {29.6} &  {29.7} & \textbf{44.3} &	 {40.9} & \textbf{40.4} \\\hline

  \end{tabular}}
  \label{tab:mainresults}
  \vspace{-10pt}
\end{table*}
\endgroup
\vspace{5pt}
\noindent\textbf{3.1.~Feature extraction module.}
\edit{The feature extraction module in \figref{fig:Ego-STAN} extracts semantically rich feature maps from ego-centric images via multiple non-linear convolutional filters. Building on a ResNet-101 \cite{he2016deep} backbone for extracting image-level features, we also introduce \emph{feature map tokens} (FMT), a specialized set of learnable parameters utilized by our Transformer to connect valuable pose information across time-steps. The FMT consists of multiple feature map points, which are formed by a weighted sum across spatial and temporal dimensions, corresponding to a particular location in an image for an intermediate 2D heatmap representation. Therefore, each unit of the FMT $\mathbf{K}$ learns how to represent accurate semantic features for the heatmap reconstruction module. By combining information from different time steps, Ego-STAN accomplishes 2D heatmap estimation even in challenging cases where views suffer from extreme occlusions.}

\noindent\textbf{3.2. Spatio-temporal Transformer.}
\edit{Self-attention learns to map the pairwise relationship between \emph{input tokens} in the sequence. This is especially important because it allows the feature map token to look across all of the input tokens in the sequence which are distributed spatially and temporally and to learn where to pay attention. As a result, feature map tokens create an accurate semantic map for heatmap reconstruction (further discussed in Sec. 3.3).} \edit{To explore the impact of FMT, we explore two variants on the spatio-temporal model without FMT. Since we are interested in estimating the 3D pose of the current frame given a sequence of frames from the past, the first variant, called \emph{slice}, takes the indices of the tokens that are respective to the current frame in the token sequence. The second variant, \emph{avg}, reduces the spatial dimensionality by taking the mean of spatially equal but temporally separated tokens.}

\noindent\textbf{3.3. Heatmap reconstruction module.}
\label{sec:heatmaprecon}
\edit{ We leverage deconvolution layers to reconstruct ground truth 2D heatmaps, $\mathbf{M}\in \mathbb{R}^{h \times w \times J}$, for each major joint ($J$) in the human body, similar to the approach of \cite{tome2019xr}.  The 2D heatmap reconstruction module, having an estimated 2D heatmap $\mathbf{\widehat{M}} \in \mathbb{R}^{h \times w \times J}$, is trained via a mean square error loss:}
\begin{equation}
    \mathcal{L}_{2D} (\mathbf{M}, \mathbf{\widehat{M}}) = \texttt{MSE}(\mathbf{M}, \mathbf{\widehat{M}}).
\label{eq:L2D}
\end{equation}

\noindent\textbf{3.4. 3D pose estimation module.}
\label{sec:hm2pose}
\edit{We leverage a simple convolution block followed by linear layers to lift the 2D heatmaps to 3D poses. As opposed to the SOTA egocentric pose estimator \cite{tome2019xr}, which uses a dual branched auto-encoder structure aimed at preserving the uncertainty information from 2D heatmaps, we (somewhat surprisingly) find that such a complex auto-encoder design is in fact not required, and our simple architecture accomplishes this task more accurately.} \edit{To estimate the 3D pose using the reconstructed 2D heatmaps \eqref{eq:L2D}, we use three different types of loss functions  -- i) squared $\ell_2$-error $\mathcal{L}_{\ell_2}(\cdot)$, ii) cosine similarity $\mathcal{L}_{\theta}(\cdot)$, and iii) $\ell_1$-error $\mathcal{L}_{\ell_1}(\cdot)$ between $\mathbf{\widehat{P}}$ and $\mathbf{P}$. These loss functions impose the closeness between  $\mathbf{P}$ and $\mathbf{\widehat{P}}$ in multiple ways. As a result, our 3D loss for regularization parameters $\lambda_{\theta}$ and $\lambda_{\ell_1}$ is}
\begin{equation}
    \mathcal{L}_{3D}(\mathbf{P}, \mathbf{\widehat{P}}) = \mathcal{L}_{\ell_2}(\mathbf{P},\mathbf{\widehat{P}}) +  \lambda_{\theta}\mathcal{L}_\theta(\mathbf{P},\mathbf{\widehat{P}})+\lambda_{\ell_1}\mathcal{L}_{\ell_1}(\mathbf{P},\mathbf{\widehat{P}})
\label{eq:L3D}
\end{equation}
\vspace*{-10pt}
\begin{equation}
    \begin{gathered}
    \mathcal{L}_{\ell_2}(\mathbf{P},\mathbf{\widehat{P}}) :=  \lVert \mathbf{\widehat{P}} - \mathbf{P} \rVert^{2}_2 , 
    \mathcal{L}_\theta(\mathbf{P},\mathbf{\widehat{P}}) :=  \textstyle\sum_{i=1}^{J} \tfrac{\langle
    \mathbf{P}_i,\mathbf{\widehat{P}}_i\rangle}{\lVert\mathbf{P}_i\rVert_2\lVert\mathbf{\widehat{P}}_i\rVert_2},\\
    \text{and~} \mathcal{L}_{\ell_1}(\mathbf{P}, \mathbf{\widehat{P}}) :=  \textstyle\sum_{i=1}^{J} \lVert \mathbf{\widehat{P}}_i - \mathbf{P}_i \rVert_1. \notag
    \end{gathered}
\end{equation}
\edit{Thus, the overall loss function to train Ego-STAN comprises the 2D heatmap reconstruction loss and the 3D loss, as shown in \eqref{eq:L2D} and \eqref{eq:L3D}, respectively.}

\section{Experiments}
\edit{We now compare the performance of Ego-STAN against the SOTA egocentric HPE methods. We consider the xR-EgoPose dataset \cite{tome2019xr}, to the best of our knowledge the only dataset which contains sequential egocentric views. We report the mean per-joint position error (MPJPE) metric.}

\tabref{tab:mainresults} shows the MPJPE achieved by Ego-STAN and its variants on the xR-EgoPose Test Set, as compared with the SOTA egocentric HPE models proposed in Tome et. al.~\cite{tome2019xr} (a dual-branch autoencoder model, and its single branch variant), and a popular outside-in baseline \cite{martinez2017simple}. Ego-STAN variants perform the best across different actions as shown in \tabref{tab:mainresults}, with Ego-STAN FMT achieving the best average performance. Ego-STAN FMT outperforms the dual-branch model proposed in \cite{tome2019xr} by a substantial \textbf{17.8~mm (30.6\%)}, averaged over all actions and joints (\tabref{tab:mainresults}).

For a qualitative comparison, we show the estimation results on a few highly self-occluded frames in \figref{fig:qualitative}, which further demonstrates the superior properties of Ego-STAN FMT over the SOTA egocentric HPE methods. Moreover, as compared to 141 million parameters in SOTA, Ego-STAN only requires 110 million parameters, resulting in a 22\% reduction while achieving significant improvement in the pose estimation performance.

\section{Conclusions}
\label{sec:discussion}

Egocentric pose estimation is challenging due to self-occlusions and distorted views. To address these challenges, we design a spatio-temporal transformer-based 3D HPE model powered by learnable parameters --- the \emph{feature map tokens} --- which can help achieve spatio-temporal attention to significantly reduce errors caused by self-occlusion. 

Our proposed Ego-STAN model(s) achieves significant performance improvements on the \xR{-EgoPose} dataset by consistently outperforming the current SOTA, while simultaneously reducing the number of trainable parameters, making it suitable for cutting-edge motion tracking applications such as activity recognition, surgical training, and immersive \xR{} applications. Our future work will investigate the role of each component via extensive ablation studies.

{\small
\bibliographystyle{ieee_fullname}
\bibliography{output}
}

\end{document}